# Ontologies and Information Extraction[*]

C. Nédellec[1] and A. Nazarenko[2]

[1] Laboratoire Mathématique, Informatique et Génome (MIG), INRA, Domaine de Vilvert, 78352 F-Jouy-en-Josas cedex

[2] Laboratoire d'Informatique de Paris-Nord (LIPN), Université Paris-Nord & CNRS, av. J.B. Clément, F-93430 Villetaneuse

## 1 Introduction

An ontology is a description of conceptual knowledge organized in a computer-based representation while information extraction (IE) is a method for analyzing texts expressing facts in natural language and extracting relevant pieces of information from these texts.

IE and ontologies are involved in two main and related tasks,
- Ontology is used for Information Extraction: IE needs ontologies as part of the understanding process for extracting the relevant information;
- Information Extraction is used for populating and enhancing the ontology: texts are useful sources of knowledge to design and enrich ontologies.

These two tasks are combined in a cyclic process: ontologies are used for interpreting the text at the right level for IE to be efficient and IE extracts new knowledge from the text, to be integrated in the ontology.

We will argue that even in the simplest cases, IE is an ontology-driven process. It is not a mere text filtering method based on simple pattern matching and keywords, because the extracted pieces of texts are interpreted with respect to a predefined partial domain model. We will show that depending on the nature and the depth of the interpretation to be done for extracting the information, more or less knowledge must be involved.

Extracting information from texts calls for lexical knowledge, grammars describing the specific syntax of the texts to be analyzed, as well as semantic and ontological knowledge. In this chapter, we will not take part in the debate about the limit between lexicon and ontology as a conceptual model. We will rather focus

---

[*] LIPN Internal Report, 2005. This paper has been originally written in march 2003. A shorter version has been published under the title "Ontology and Information Extraction: A Necessary Symbiosis", in *Ontology Learning from Text: Methods, Evaluation and Applications*, edited by P. Buitelaar, P. Cimiano and B. Magnini, *IOS Press Publication*, July 2005,

on the role that ontologies viewed as *semantic knowledge bases* could play in IE. The ontologies that can be used for and enriched by IE relate conceptual knowledge to its linguistic realizations (*e.g.* a concept must be associated with the terms that express it, eventually in various languages).

Interpreting text factual information also calls for knowledge on the domain referential entities that we consider as part of the ontology (Sect. 2.2.1).

This chapter will be mainly illustrated in biology, a domain in which there are critical needs for content-based exploration of the scientific literature and which becomes a major application domain for IE.

## 2 Settings

Before exploring the close relationship that links ontology and IE in Sect. 3 and Sect. 4, we will define Information Extraction and ontology.

### 2.1 What is IE?

The considerable development of multimedia communication goes along with an exponentially increasing volume of textual information. Today, mere Information Retrieval (IR) technologies are unable to meet the needs of specific information because they provide information at a document collection level. Developing intelligent tools and methods, which give access to document content and extract relevant information, is more than ever a key issue for knowledge and information management. IE is one of the main research fields that attempt to fulfill this need.

#### 2.1.1 Definition

The IE field has been initiated by the DARPA's MUC program (Message Understanding Conference in 1987 (MUC Proceedings; Grishman and Sundheim 1996). MUC has originally defined IE as the task of (1) extracting specific, well-defined types of information from the text of homogeneous sets of documents in restricted domains and (2) filling pre-defined form slots or templates with the extracted information. MUC has also brought about a new evaluation paradigm: comparing the information extracted by automatic ways to human-produced results. MUC has inspired a large amount of work in IE and has become a major reference in the text-mining field. Even as such, it is still a challenging task to build an efficient IE system with good recall (coverage) and precision (correctness) rates.

A typical IE task is illustrated by Fig. 1 from a CMU corpus of seminar announcements (Freitag 1998). IE process recognizes a name (John Skvoretz) and classifies it as a person name. It also recognizes a seminar event and creates a seminar event form (John Skvoretz is the seminar speaker whose presentation is entitled "Embedded commitment").

Even in such a simple example, IE should not be considered as a mere keyword filtering method. Filling a form with some extracted words and textual fragments involves a part of interpretation. Any fragment must be interpreted with respect to its "context" (*i.e.* domain knowledge or other pieces of information extracted from the same document) and according to its "type" (*i.e.* the information is the value of an attribute / feature / role represented by a slot of the form). In the document of Fig. 1, "4-5:30" is understood as a time interval and background knowledge about seminars is necessary to interpret "4" as "4 pm" and as the seminar starting time.

> **Form to fill** (partial)
>     place:?
>     starting time: ?
>     title: ?
>     speaker: ?
> **Document**: Professor John Skvoretz, U. of South Carolina, Columbia, will present a seminar entitled "Embedded commitment", on Thursday, May 4th from 4-5:30 in PH 223D.
> **Filled form** (partial)
>     place: PH 223D
>     starting time: 4 pm
>     title: Embedded commitment
>     speaker: Professor John Skvoretz […]

**Fig 1.** A seminar announcement event example

### *2.1.2 IE overall process*

Operationally, IE relies on document preprocessing and extraction rules (or extraction patterns) to identify and interpret the information to be extracted. The extraction rules specify the conditions that the preprocessed text must verify and how the relevant textual fragments can be interpreted to fill the forms. In the simplest case, the textual fragment and the coded information are the same and there are neither text preprocessing nor interpretation.

More precisely, in a typical IE system, three processing steps can be identified (Hobbs *et al.* 1997; Cowie and Wilks 2000):

1. *text preprocessing,* whose level ranges from mere text segmentation into sentences and sentences into tokens to a full linguistic analysis;
2. *rule selection*: the extraction rules are associated with triggers (*e.g.* keywords), the text is scanned to identify the triggering items and the corresponding rules are selected;
3. *rule application,* which checks the conditions of the selected rules and fills the forms according to the conclusions of the matching rules.

**Extraction rules.** The rules are usually declarative. The conditions are expressed in a Logics-based formalism (Fig. 3), in the form of regular expressions, patterns or transducers. The conclusion explains how to identify in the text the value that should fill a slot of the form. The result may be a filled form, as in Fig. 1 and 2, or equivalently, a labeled text as in Fig. 3.

```
Sentence: "GerE stimulates the expression of cotA."
Rule
    Conditions: X="expression of"
    Conclusions: Interaction_Target <-next-token(X).
Filled form: Interaction_Target: cotA
```

**Fig. 2.** IE partial example from functional genomics

Experiments have been made with various kinds of rules, ranging from the simplest ones (Riloff 1993) (*e.g.* the subject of the passive form of the verb "murder" is interpreted as a victim) to sophisticated ones as in (Soderland *et al.* 1995). The more explicit (*i.e.* the more semantic and conceptual) the IE rule, the more powerful, concise and understandable it is. However, it requires the input text being parsed and semantically tagged.

A single slot rule extracts a single value, as in Fig. 2, while a multi-slot rule correctly extracts at the same time all the values for a given form as in Fig. 3, even if there is more than one event reported in the text fragment.

**IE forms.** Extraction usually proceeds by filling forms of increasing complexity (Wilks 1997):

- *Filling entity forms* aims at identifying the items representing the domain referential entities. These items are called "named entities" (*e.g. Analysis & Technology Inc.*) and assimilated to proper names (company, person, gene names) but they can be any kind of word or expression that refers to a domain entity: dates, numbers, titles for the management succession MUC-6 application, bedrooms in a real-estate IE application (Soderland 1999).
- *Filling domain event forms*: The information about the events extracted by the rules is then encoded into forms in which a specific event of a given type and its role fillers are described. An entity form may fill an event role.
- *Merging forms* that are issued from different parts of the text but provide information about a same entity or event.
- *Assembling scenario forms*: Ideally, various event and entity forms can be further organized into a larger scenario form describing a temporal or logical sequence of actions/events.

**Text processing.** As shown in Fig. 3, the condition part of the extraction rules may check the presence of a given lexical item (*e.g.* the verb *named*), the syntactic category of words and their syntactic dependencies (*e.g.* object and subject relations). Different clues such as typographical characteristics, relative position of words, semantic tags[1] or even coreference relations can also be exploited.

Most IE systems therefore involve linguistic text processing and semantic knowledge: segmentation into words, morpho-syntactic tagging (the part-of-speech categories of words are identified), syntactic analysis (sentence constituents such as noun or verb phrases are identified and the structure of complex sen-tences is analyzed) and sometimes additional processing: lexical disambiguation, semantic tagging or anaphora resolution.

---

[1] *E.g.*, if the verbs "named", "appointed" and "elected" of Fig.3 were all known as 'nomination' verbs, the fourth condition of the rule could have been generalized to their semantic category 'nomination'.

> **Sentence**: "NORTH STONINGTON, Connecticut (Business Wire) - 12/2/94 - Joseph M. Marino and Richard P. Mitchell have been named senior vice president of Analysis & Technology Inc. (NASDAQ NMS: AATI), Gary P. Bennett, president and CEO, has announced. "
> **Rule**
>   *Conditions*:
>   noun-phrase (PNP, head(isa(person-name))),
>   noun-phrase (TNP, head(isa(title))),
>   noun-phrase (CNP, head(isa(company-name))),
>   verb-phrase (VP, type(passive),head(named or elected or appointed)),
>   preposition (PREP, head(of or at or by)),
>   subject (PNP, VP),
>   object (VP, TNP),
>   post_nominal_prep (TNG,PREP),
>   prep_object (PREP, CNP)
>   *Conclusion*:
>   management_appointment (M, person(PNP), title (TNP), company (CNP)).
>   *Comment*:
> **if** there is a noun phrase (NP) whose head is a person name (PNP), an NP whose head is a title name (TNP), an NP whose head is a company name (CNP), a verb phrase whose head is a passive verb (*named* or *elected* or *appointed*), a preposition *of*, *at* or *by*,
> **if** PNP and TNP are respectively subject and object of the verb,
> **and if** CNP modifies TNP,
> **then** it can be stated that the person "PNP" is named "TNP" of the company "CNP".
> **Labeled document**
> NORTH STONINGTON, Connecticut (Business Wire) - 12/2/94 - <Person>Joseph M. Marino and Richard P. Mitchell</Person> have been named <Title>senior vice president</Title> of <Company>Analysis & Technology Inc</Company>. (NASDAQ NMS: AATI), Gary P. Bennett, president and CEO, has announced.

**Fig. 3.** Example from MUC-6, a newswire about management succession

However, the role and the scope of this analysis differ from one IE system to another. Text analysis can be performed either as preprocessing or during extraction rule application. In the former case, the whole text is first analyzed. The analysis is global in the sense that items spread all over the document can contribute to built the normalized and enriched representation of the text. Then, the application of extraction rules comes to a simple filtering process of the enriched representation. In the latter case, the text analysis is driven by the rule condition verification. The analysis is local, focuses on the context of the triggering items of the rules, and fully depends on the conditions to be checked in the selected rules.

In the first IE systems (Hobbs *et al.* 1997), local and goal-driven analysis was preferred to full text preanalysis to increase efficiency, and the text preprocessing step was kept to minimum. Although costly, data-driven, full text analysis and normalization can improve the IE process in various manners. (1) It improves further NL processing steps, *e.g.* syntactic parsing improves attachment disambiguation (Basili *et al.* 1993) or coreference resolution. (2) Full text analysis and normalization also facilitates the discovery of lexical and linguistic regularities in specific documents. This idea, initially promoted by works on sublanguages (Har-ris 1968, Sager *et al.* 1987) for tuning NL processing to a given type of texts, is now popularized by Machine Learning (ML) papers in the IE field for learning extraction rules. There are two main reasons for that. First, annotating training data is costly and the quantity of data to be annotated decreases with the normalization (the less variations in the data, the less data annotation is needed). Next, ML systems tend to learn non-understandable rules by picking details in training examples that do not look as related. Normalizing the text by representing it in a more abstract way increases the understandability of the learned rules. However, normalization also raises problems such as the biased choice of the right representation *before learning*, that is not dealt with in the IE literature.

We will see in the following that these two approaches, in which text analysis is respectively used for interpretation (goal-driven) and normalization (data-driven), are very much tangled, as any normalization process involves a part of interpretation. One of the difficulties in designing IE systems is to set the limit between local and global analysis. Syntactic analysis or entity recognition can be performed on a local basis but are improved by knowledge inferred at a global level. Thus, ambiguous cases of syntactic attachments or entity classification can be solved by comparison with non-ambiguous similar cases of the same document.

### *2.1.3 IE, an ambitious approach to text exploration*

As mentioned above, there is a need for tools that give a real access to the document content. IE and Question Answering (Q/A) tasks both try to identify in documents the pieces of information that are relevant to a given query. They differ, however, in the type of information that is looked for. A Q/A system has to answer to a wide range of unpredictable user questions. In IE, the information that is looked for is richer but the type of information is known in advance. The relevant pieces of text have to be identified and then interpreted with respect to the knowledge partially represented in the forms to fill.

IE and Q/A systems both differ in their empirism from their common ancestors, the text-understanding systems. They both rely on targeted and local techniques of text exploration rather than on a large coverage and in-depth semantic analysis of the text. The MUC competition framework has gathered a large and stable IE community. It has also drawn the research towards easily implementable and efficient methods rather than strong and well-founded NLP theories.

The role of semantics in IE is often reduced to very shallow semantic labeling. Semantic analysis is rather considered as a way to disambiguate syntactic steps than as a way to build a conceptual interpretation. Today, most of the IE systems that involve semantic analysis exploit the most simple part of the whole spectrum of domain and task knowledge, that is to say, named entities. However, the growing need for IE application to domains such as functional genomics that require more text understanding pushes towards more sophisticated semantic knowledge resources and thus towards ontologies viewed as conceptual models, as it will be shown in this chapter.

## 2.2 What is an Ontology in the IE framework?

Even though ontologies usually do not appear as an autonomous component or resource in IE systems, we argue that IE relies on ontological knowledge.

### 2.2.1 Ontologies populated with referential entities

The ontology identifies the entities that have a form of existence in a given domain and specifies their essential properties. It does not describe the spurious properties of these entities. On the contrary, the goal of IE is to extract factual knowledge to instantiate one or several predefined forms. The structure of the form (*e.g.* Fig. 4) is a matter of ontology whereas the values of the filled template usually reflect factual knowledge (as shown in Fig. 2 above) that is not part of the ontology. In these examples, the form to fill represents a part of the biological model of gene regulation network: proteins interact positively or negatively with genes. In Sect. 3.4, we will show that IE is ontology-driven in that respect.

| | |
|---|---|
| **Interaction** | **Type:** {negative, positive} |
| | **Agent**: *any protein* |
| | **Target**: *any gene* |

**Fig. 4.** An example of IE form in the genomics domain

The status of the named entities is a pending question. Do they belong to the ontology or are they factual knowledge? From a theoretical point of view, according to Brachman's terminological logics view (1979), they are instances of concepts and as such, they are described and typed at the *assertional* level and not at the terminological or ontological level. In this chapter, we will nevertheless consider that entities, being *referential* entities, are part of the domain ontology because it is the way IE considers them.

### 2.2.2 Ontology with a natural language anchorage

Whether one wants to use ontological knowledge to interpret natural language or to exploit written documents to create or update ontologies, in any case, the ontology has to be connected to linguistic phenomena. Ontology must be linguistically anchored. A large effort has been devoted in traditional IE systems based on local analysis to the definitions of extraction rules that achieve this anchoring. In the very simple example about gene interaction (Fig. 2 above), the ontological knowledge is encoded as a keyword rule, which can be considered as a kind of compiled knowledge. In more powerful IE systems, the ontological knowledge is more explicitly stated in the rules that bridge the gap between the word level and text interpretation. For instance, the rule of Fig. 3 above, states that a management appointment event can be expressed through three verbs (*named*, *elected* or *appointed*). As such, an ontology is not a purely conceptual model, it is a model associated to a domain-specific vocabulary and grammar. In the IE framework, we consider that this vocabulary and grammar are part of the ontology, even when they are embodied in extraction rules.

The complexity of the linguistic anchoring of ontological knowledge is well known and should not be underestimated. A concept can be expressed by different terms and many words are ambiguous. Rhetoric, such as lexicalized metonymies or elisions, introduces conceptual shortcuts at the linguistic level and must be elicited to be interpreted into domain knowledge. A noun phrase (*e.g.* "the citizen") may refer to an instance (a specific citizen which has been previously mentioned in the text) or to the class (the set of all the citizens) leading then to a very different interpretation. These phenomena, which illustrate the gab between the linguistic and the ontological levels, strongly affect IE performance. This explains why IE rules are so difficult to design.

### *2.2.3 Partial ontologies*

IE is a targeted textual analysis process. The target information is described in the structure of the forms to fill. As mentioned above (Sect. 2.1.2) MUC has identified various types of forms describing elements or entities, events and scenarios.

IE does not require a whole formal ontological system but parts of it only. We consider that the ontological knowledge involved in IE can be viewed as a set of interconnected and concept-centered descriptions, or "conceptual nodes[2]". In conceptual nodes the concept properties and the relations between concepts are explicit. These conceptual nodes should be understood as chunks of a global knowledge model of the domain. We consider here various types of concepts: an *object node* lists the various properties of the object; an *event node* describes the various objects involved in the event and their roles; a *scenario node* describes one or several events involved in the scenario and their interrelations. The use of this type of knowledge in NLP systems is traditional (Schank and Abelson 1977) and is illustrated by MUC tasks.

### **2.3 Specificity of the ontology-IE relationship**

Ontology and IE are closely connected by a mutual contribution. The ontology is required for the IE interpreting process and IE provides methods for ontological knowledge acquisition. Even if using IE for extracting ontological knowledge is still rather marginal, it is gaining in importance. We distinguish both aspects in the following Sects. 3 and 4, although the whole process is a cyclic one. A first level of ontological knowledge (*e.g.* entities) helps to extract new pieces of knowledge from which more elaborated abstract ontological knowledge can be designed, which help to extract new pieces of information in an iterative process.

---

[2] We define a conceptual node as a piece of ontological model to which linguistic information can be attached. It differs from the "conceptual nodes" of (Soderland *et al*. 1995), which are extraction patterns describing a concept. We will see below that several extraction rules may be associated to a unique conceptual node.

## 3. Ontology for Information extraction

The template or form to be fulfilled by IE is a partial model of world knowledge. IE forms are also classically viewed as a model of a database to be filled by the instances extracted. This view is consistent with the first one. In this respect, any IE system is ontology-driven: in IE processes, the ontological knowledge is primarily used for text interpretation. How poor the semantics underlying the form to fill may be (see Fig. 2, for instance), whether it is explicit (Gaizauskas and Wilks, 1997; Embley *et al.*, 1998) or not (Freitag 1998) (see Fig. 5 below), IE is always based on a knowledge model. In this Sect. 3, for exposition purposes, we distinguish different levels of ontological knowledge:

- The referential domain entities and their variations are listed in "flat ontologies". This is mainly used for entity identification and semantic tagging of character strings in documents.
- At a second level, the conceptual hierarchy improves normalization by enabling more general levels of representation.
- More sophisticated IE systems also make use of chunks of a domain model (*i.e.* conceptual nodes), in which the properties and interrelations of entities are described. The projection of these relations on the text both improves the NL processes and guides the instantiation of conceptual frames, scenarios or database tuples. The corresponding rules are based either on lexico-syntactic patterns or on more semantic ones.
- The domain model itself is used for inference. It enables different structures to be merged and the implicit information to be brought to light.

### 3.1 Sets of entities

Recognizing and classifying named entities in texts require knowledge on the domain entities. Specialized lexical or key-word lists are commonly used to identify the referential entities in documents. For instance, in the context of cancer treatment, (Rindflesh *et al.* 2000) makes use of the concepts of the Metathesaurus of UMLS to identify and classify biological entities in papers reporting interactions between proteins, genes and drugs. In different experiments, some lists of gene and protein names are exploited. For instance, (Humphreys *et al.* 2000) makes use of the SWISS PROT resource whereas (Ono *et al.* 2001) combines pattern matching with a manually constructed dictionary. In the financial news of MUC-5, lists of company names have been used. In a similar way, Auto-Slog (Riloff 1993), CRYSTAL (Soderland *et al.* 1995), PALKA (Kim and Moldovan 1995), WHISK (Soderland 1999) and Pinocchio (Ciravegna 2000) make use of list of entities to identify the referential entities in documents. The use of lexicon and dictionaries is however controversial. Some authors like (Mikheev *et al.* 1999) argue that entity named recognition can be done without it.

Three main objectives of these specialized lexicons can be distinguished, semantic tagging, naming normalization and linguistic normalization, although these operations are usually processed all at once.

*Semantic tagging*

**Semantic tagging.** List of entities are used to tag the text entities with the relevant semantic information. In the ontology or lexicon, an entity (*e.g.* Tony Bridge) is described by its type (the semantic class to which it belongs, here PERSON) and by the list of the various textual forms (typographical variants, abbreviations, synonyms) that may refer to it[3] (*Mr. Bridge, Tony Bridge, T. Bridge*).

However, exact character strings are often not reliable enough for a precise entity identification and semantic tagging. Polysemic words that do exist even in sublanguages belong to different semantic classes. In the above example, the string "Bridge" could also refer to a bridge named "Tony". (Soderland 1999) reports experiments on a similar problem on a software job ad domain: WHISK is able to learn some contextual IE rules but some rules are difficult to learn because they rely on subtle semantic variations, *e.g.*, the word "Java" can be interpreted as competency in the programming language except in "Java Beans". Providing the system with lists of entities does not help that much, "because too many of the relevant terms in the domain undergo shifts of meaning depending on context for simple lists of words to be useful". The connection between the ontological and the textual levels must therefore be stronger. Identification and disambiguation contextual rules can be attached to named entities.

This disambiguation problem is addressed as an autonomous process in IE works by systems that learn contextual rules for entity identification (Sect. 4.1).

**Naming normalization.** As a by-effect, these resources are also used for normalization purposes. For instance, the various forms of Mr. Bridge will be tagged as MAN and associated with its canonical name form: Tony Bridge (<PERSON id=Tony Bridge>). In (Soderland 1999), the extraction rules may refer to some class of typographical variations (such as `Bdrm=(brs, br, bdrm, bedrooms, bedroom, bed)` in the Rental Ad domain). This avoids rule overfitting by enabling then specific rules to be abstracted.

Specialized genomics systems are particularly concerned with the variation problem, as the nomenclatures are often not respected in the genomics literature, when they exist. Thus, the well-known problem of identifying protein and gene names has attracted a large part of the research effort in IE to genomics (Proux *et al.* 1998; Fukuda *et al.* 1998; Collier *et al.* 2000). In many cases, rules rely on shallow constraints rather than morpho-syntactic dependencies.

**Linguistic normalization.** Beyond typographical normalization, the semantic tagging of entities contributes to sentence normalization at a linguistic level. It solves some syntactic ambiguities, *e.g.* if *cotA* is tagged as a *gene*, in the sentence "the stimulation of the expression of cotA", knowning that a gene can be "expressed" helps to understand that "cotA" is the patient of the expression rather than its agent or the agent of the stimulating action. Semantic tagging is also traditionally used for anaphora resolution: (Pustejovsky *et al.* 2002) makes use of UMLS types to identify and order the potential antecedents of an anaphoric pronoun (*it*) or noun phrase (*these enzymes*, *both genes*).

---
[3] These various forms may be listed extensionally or intentionally by variation rules.

Semantic types and naming variations are both used for text normalization, without a clear distinction between them.

### 3.2 Hierarchies

Beyond lists of entities, ontologies are often described as hierarchies of semantic or word classes. Traditionally, IE focuses on the use of word classes rather than on the use of the hierarchical organization. For instance, in WordNet (Miller 1990), the word classes (*synsets*) are used for the semantic tagging and disambiguation of words but the hyponymy relation that structures the synsets into a hierarchy of semantic or conceptual classes is seldom exploited for ontological generalization inference. Some ML-based experiments have been done to exploit hierarchies of WordNet and of more specific lexicons, such as UMLS (Soderland *et al.* 1995; Chai *et al.* 1999; Freitag 1998). The ML systems learn extraction rules by generalizing from annotated training examples. They relax constraints along two axes, climbing the hyperonym path and dropping conditions. This way, the difficult choice of the correct level in the hierarchy is left to the systems.

(Chai *et al.* 1999) reports experiments that show how difficult is this choice in WordNet. Their IE patterns are in the form of pairs of noun phrases (NP) representing two target values, related by a verb or a preposition representing the relation, such as `NP(X->Enzyme) Verb(interact) NP(Y->Gene)` where `Enzyme` and `Gene` represent two slots of a form. In this example, the categories of `X` and `Y` are not constrained and the pattern is over general. As one may have expected, the experiment shows that generalizing the two NP types with WordNet increases the recall but decreases the precision. Chai *et al.* system automatically learns for each relevant NP in the pattern, the optimal level of semantic generalization on the WordNet hyperonym path by climbing WordNet hierarchies. For ambiguous words, which have several hyperonyms, the choice of the right hierarchy to climb is based on the user selection of the headword senses in a training corpus. The rule learned using WordNet enhances the overall F-measurement (combination of precision and recall) by about 10 %. The performance increases by about 30 % for certain facts such as LOCATION in a job advertisement IE task. The conclusion is moderate: generalization along WordNet hierarchy brings a significant benefit to IE but the incompleteness of WordNet in specific domain and the word sense ambiguity are questionable.

```
    acqabr :-
    some(Var, null, capitalized, true), length(>2)
    some(Var [next_token], all_lower_case, true),
    some(Var [prev_token], all_lower_case, true),
    some(Var [right_AN], wordnet_word, "possession"),
    some(Var [right_AN prev_token], wordnet_word, "stock"),
    some(Var [prev_tok prev_token], doubleton, false).
applies to the sentences,
    "to purchase 4.5 mln Trilogy common shares at"
    "acquire another 2.4 mln Roach treasury shares."
where Possession (instanciated by shares) and stock (instanciated by common and
treasury) are WordNet classes. Var represents the company name (Trilogy and
Roach) and Right_AN represents the head of the noun phrase at the right of the
company name.
```

**Fig. 5.** Application of SRV to MUC-5

The IE learning system, SRV, also uses semantic class information such as synsets and hyperonym links from WordNet lexicon to constrain the application of the IE rules (Fig. 5), but D. Freitag (1998) concludes that the improvement is not clear.

A lot of work has been devoted to the manual or automatic acquisition of domain dependent hierarchies for a wide range of NL processing tasks in order to overcome the general ontologies limitations. For instance, for IE purpose, (Riloff and Shepherd 1997) proposes to build semantic classes starting with a few seed words and growing by adjunction of words as a first step before extraction rules learning. At this stage, no clear and reusable conclusions for IE can be drawn from these attempts.

### 3.3 Conceptual nodes

The ontological knowledge is not always explicitly stated as it is in (Gaizauskas and Wilks 1997), which represents an ontology as a hierarchy of concepts, each concept being associated with an attribute-value structure, or in (Embley *et al.* 1998), which describes an ontology as database relational schema. However, ontological knowledge is reflected by the target form that IE must fill and which represents *the conceptual nodes* to be instantiated. Extraction rules ensure the mapping between a conceptual node and the potentially various linguistic phrasing expressing the relevant elements of information.

Most of the works aiming at extracting gene/protein interactions are based on such event conceptual nodes. In (Yakushiji *et al.* 2001), predicate-argument structures (P-A structures), also referred as subcategorization frames, describe the number, type and syntactic construction of the predicate arguments. The P-A structures are used for extracting gene and protein interactions (see Fig. 7). The mapping between P-A structures and event frames (event conceptual nodes) is explicit and different P-A structures can be associated to a same event frame. For instance, the extraction of gene/protein interactions is viewed as the search for the subject and the object of an interaction verb, which are interpreted as the agent and the target of the interaction.

In these works, parsing is made by shallow, robust or full parsers, which handle or not coordinates, anaphora, passive mood and nominalization (Sekimizu *et al.* 1998; Thomas *et al.* 2000; Roux *et al.* 2000; Park *et al.* 2001; Leroy and Chen 2002). Additional semantic constraints may be added as selectional restrictions[4] for disambiguation purposes.

```
activate is an interaction verb
P-A structure of activate:
Pred: activate                    Frame: activate
    args:    subject (1)              slot:    agent (1)
             object (2)               slot:    target (2)
```

**Fig 6.** Example of a conceptual-node driven rule in functional genomics

These approaches rely on the assumption that semantic relations (*e.g.* agent, target) are fully determined by the verb/noun predicate, its syntactic dependencies and optionally the semantic categories of its arguments, (Pustejovsky *et al.* 1993; Gildea and Jurafsky 2002).

The same assumption is made in the very interesting work of (Sasaki and Matsuo 2000) that goes one step further. Their system, RHB+, learns this mapping with the help of case-frames in Fillmore's sense (1968). RHB+ is able to combine multiple case-frames to map a unique conceptual node, as opposed to the direct binary mapping described above. RHB+ makes use of Japanese linguistic resources that include a 12-level hierarchical concept thesaurus of 3,000 categories (300,000 words) linked by is-a and has-a relations and 15,000 case frames for 6,000 verbs. The case-roles, the semantic relations between a predicate and its argument, within a text are determined by the semantic categories of the predicate arguments together with the prepositions or the syntactic dependencies.

As for RHB+, considerable effort has been made towards designing automatic methods for learning such extraction rules. The main difficulty arises from the complexity of the text representation once enriched by the multiple linguistic and conceptual levels. The more expressive the representation, the larger is the search space for the IE rule and the more difficult the learning. The extreme alternative consists in either selecting the potentially relevant features before learning with the risk of excluding the solution from the search space, or leaving the system the entire choice, provided that there is enough representative and annotated data to find the relevant regularities. For instance, the former consists in normalizing by replacing names by category labels when the latter consists in tagging without removing the names. The learning complexity can even be increased when the conceptual or semantic classes are learned together with the conceptual node information (Riloff and Jones 1999; Yangarber *et al.* 2000).

---

[4] A selectional restriction is a semantic type constraint that a given predicate enforces on its arguments.

An interesting alternative to the purely corpus-based approach for learning IE rules has been proposed in the context of ECRAN (Basili *et al.* 1998). This approach, based on Wilks's notion of lexical tuning, consists in adapting an existing dictionary to a given corpus, the LDOCE dictionary, which aims at describing the subcategorization frames of any word sense. The confrontation of a specialized corpus syntactic analysis and the LDOCE allows the selection of the most relevant subcategorization frames and possibly the discovery of new word senses.

### 3.4 Domain conceptual model

The link between the syntactic level and the event description is not always so straightforward. The text interpretation may require inference reasoning with domain knowledge. For instance, to be able to extract

| Interaction | Type: negative |
| --- | --- |
| | Agent: sigma K |
| | Target: spoIIID |

from, "[…], such that production of *sigma K* leads to a decrease in the level of *spoIIID*.", more biological knowledge is necessary to interpret the level changes in term of interaction. P-A structures as those above will be useful at the lower level for interpreting the text and build a semantic structure but a causal model stating that correlation in quantity variations can be interpreted as an interaction is needed to connect and interpret the instantiated syntactic structures at a conceptual level.

## 4 Information extraction for ontology design

Acquisition of ontological knowledge is a well-known bottleneck for many AI applications and a large amount of work has been devoted to knowledge acquisition from text. The underlying idea, inherited from Harris' work on the immunology sublanguage (Harris *et al.* 1989), is that, in specific domains, the linguistics reflects the domain conceptual organization. Although it has been observed, as we mentioned above, that the linguistic representation of the conceptual domain is biased, it remains one of the most promising approaches to knowledge acquisition. Following (Meyer *et al*. 1992), a large amount of work has been devoted to term extraction (Bourigault 1996, Jacquemin 1996) as a mean to identify the concepts of a given domain and thus to bootstrap ontology design (Grishman and Sterling 1992; Nazarenko *et al*. 1997; Aussenac-Gilles *et al.* 2000). Identifying how these terms relate to each other in texts help to understand the properties and relationships of the underlying concepts.

Various methods are applied to corpora to achieve this acquisition process: endogenous distributional or cooccurrence analysis and rule-based extraction are complementary in this respect. We focus here on the latter approach, which pertains to IE. We show that it can indeed contribute to the ontology acquisition and enrichment process. Rule-based extraction produces elementary results that are interpreted in terms of chunks of ontological knowledge: the referential entities and their interrelationships. Once extracted, these chunks have to be integrated into the ontology. We do not deal with that point here, as it goes beyond IE.

**4.1 Entity name extraction**

As explained in Sect. 2.2, we consider here that the referential entities (*e.g.* persons, dates or genes), which are usually represented as instances of concepts, are part of the ontology. In this perspective, there is a need for "populating" the ontology with the referential entities of the domain of interest by automatic ways.

IE has been widely applied to the recognition and categorization of the entities mentioned in documents, either specialized texts or web pages by means of patterns. The extraction methods differ regarding their pattern design technique, which is either automatic or manual.

*4.1.1 Automatic pattern learning*

Hidden Markov Models (HMM) based on sequences of bigrams (pairs of tokens) has become a popular method for learning named entity recognition patterns from annotated corpora since Nymble (Bikel *et al.* 1997, 1999) because simple bigrams appear as sufficient for learning efficient rules. According to (Collier *et al.* 2000), the applied HMM differ in their ability to learn the model structure or not, in the way they estimate the transition probabilities (from training data or models built by hand) and in their reusability in different domains. For instance, the method of (Collier *et al.* 2000) aims at recognizing biological entity names and is based on an HMM trained on 100 MedLine abstracts using only character features and lexical information. The results (F-score 73 %) are much better than those obtained by previous hand-coded patterns (Fukuda *et al.* 1998).

*4.1.2 Hand-coded patterns and dictionaries*

While the pattern learning approach tends to use very basic information from the text, the hand-coded pattern approach on contrary rely more on linguistics (Proux *et al.* 1998), external ontologies (Rindflesh *et al.* 2000) and context (Humphreys *et al.* 2000; Fukuda *et al.* 1998; Hishiki *et al.* 1998).

The EDGAR system (Rindflesh *et al.* 2000) identifies unknown gene names and cell lines by two ways: the concepts of UMLS and hand-coded contextual patterns, such as appositives, filtered through UMLS and an English dictionary and occurring after some signal words, (*e.g.* cell, clone and line for cells). A second phase identifies cell features, (*e.g.* organ type, cancer type and organism) by a similar mechanism.

(Hishiki *et al.* 1998) gives examples of contextual regular expressions applied to the entity recognition and categorization. They rely, for instance, on:
- Indefinite appositions: the pattern `NP(X), a NP(Y)` gives X as an instance of Y, if Y is a type. From the sentence "csbB, a putative membrane-bound glucosyl transferase", csbB is interpreted as an instance of transferase if transferase is defined as a type.
- Exemplification of copula constructions: `NP(X) be one of NP(Y)` or `NP(X) e.g. NP(Y)`. The fact that abrB is an instance of gene is extracted from "to repress certain genes , *e*.g.. abrB".

The recognition of gene names (Proux *et al.* 1998) and biological entity names (Humpreys *et al.* 2000) can also rely on various linguistic-based methods, i.e. grammatical tagging, contextual hand-coded patterns, specific lexicon (*e.g.* SWISS-PROT keywlist) and combination of morphology-based, biochemical suffix and prefix recognition rules. The results obtained by (Proux *et al.* 1998) on a FlyBase corpus are of high quality, (94,4 % recall and 91,4 % precision). With comparable performance, (Humpreys *et al.* 2000) identifies 25,000 component terms of 52 categories as MUC named entity results. Populating ontology with help of entity name recognition from textual data can therefore be considered as operational for specific domains.

## 4.2 Relation extraction

In a structured ontology, the concepts are related to each other according to a variety of relations. Three main approaches acquire ontological relations from texts:
- The cooccurrence-based method identifies couples of cooccurring terms. When applied to large corpora, this method is robust but further interpretation is required to type the relation underlying the collocation.
- The knowledge-based method makes use of a bootstrapping dictionary, a thesaurus or an ontology and tunes it to adapt it to the specific domain at hand according to a representative "tuning" corpus.
- The IE pattern-based method.

The IE approach has the advantage over the first one that the type of extracted relation is known, since patterns are designed to characterize a given relation. It is complementary to the second one: preexisting knowledge can help to design an extraction rule in an acquisition iterative process. For instance, if the preexisting knowledge base states that 'X is-part-of Y', identifying this relation in text helps to design a first is-part-of extraction rule, which is used in turn to extracts new instances of the that relation (Hearst 1992; Morin and Jacquemin 1999).

Two kinds of relations can roughly be distinguished: the generic ones, which can be found in almost any ontology, and the model-specific ones.

### *4.2.1 Generic relations*

The links that form the main structure of the ontology are the most popular relations: the intra-concept relations (synonymy) and the hierarchical *is-a* and *part-of* relations. They can be considered either at the linguistic level (hyperonymy and meronymy are traditional lexicographic relations) or at the ontological level (is-a and part-of). The acquisition goal is to exploit the linguistic organization to infer an ontological structure.

**Hyponymy or is-a relation.** In her pioneering work, M. Hearst (1992) proposes six patterns for the acquisition of hyponyms, among which are the following:
```
Such NP as {NP,}* {or|and} NP
NP {,} including {NP,}* {or|and} NP
```
The first NP is interpreted as the hyperonym, and the latter one(s) as the hyponym(s). The first rule matches the sentence "… such exotic fruits as kiwis, mangoes, pineapple or coconuts…" from which the relations *kiwi is-a exotic fruit* and *mango is-a exotic fruit* can be extracted. These patterns are in the form of regular expressions that combine lexical units, punctuation marks and morpho-syntactic tags. It requires a tagged corpus.

Many works have followed this track. Variant forms of exemplification and enumeration patterns have been designed for specific corpora (see the above patterns proposed by (Hishiki *et al.* 1998)). The results have obviously to be validated by a human expert and a taxonomy can be then constructed by inference on the single types derived from the corpus by pattern matching.

As shown on the previous examples, the hyponym relation can be interpreted either as an instance-class relation or as a generalization relation between two classes. The language does not distinguish the one from the other, since a prototypical instance may refer to the class as a whole. For example, from "PP2C family of eukaryotic Ser/Thr protein phosphatases", one can derive two relations (PP2C is a phosphatase and phosphatase is a protein), where classes and instances cannot be distinguished.

As one may expect, this approach gives high precision scores but no reliable measure of recall, which would call for a corpus where hyponyms are tagged. The number of extracted relations seems to be low regarding the diversity of corpus information as well as the number of ontological categories. It is generally agreed that ontology design imposes to favor reliability over cover. The obvious `NP is a NP` pattern is usually disregarded as too imprecise. In the sentence "the key regulator is an example of…" it would lead to interpret the key regulator as an instance of example.

**Meronymy or part-of relation.** Meronymy has not been as much studied as hyponymy. However, (Berland and Charniak 1997) adapts the above approach to find parts of physical objects in very large corpora. Their acquisition method is designed for ontology enhancement. They propose five patterns such as
```
Such NP as {NP,}* {or|and} NP
NP {,} including {NP,}* {or|and} NP
```
The first NP is interpreted as the hyperonym, and the latter one(s) as the hyponym(s). The first rule matches the sentence "… such exotic fruits as kiwis, mangoes, pineapple or coconuts…" from which the relations *kiwi is-a exotic fruit* and *mango is-a exotic fruit* can be extracted. These patterns are in the form of regular expressions that combine lexical units, punctuation marks and morpho-syntactic tags. It requires a tagged corpus.

Many works have followed this track. Variant forms of exemplification and enumeration patterns have been designed for specific corpora (cf. the above patterns proposed by (Hishiki *et al.* 1998)). The results have obviously to be validated by a human expert and a taxonomy can be then constructed by inference on the single types derived from the corpus by pattern matching.

As shown by the previous examples, the hyponym relation can be interpreted either as an instance-class relation or as a generalization relation between two classes. The language does not distinguish the one from the other, since a prototypical instance may refer to the class as a whole. For example, from "PP2C family of eukaryotic Ser/Thr protein phosphatases", one can derive two relations (PP2C is a phosphatase and phosphatase is a protein), where classes and instances cannot be distinguished.

As one may expect, this approach gives high precision scores but no reliable measure of recall, as it would call for a corpus where hyponyms are tagged. The number of extracted relations seems to be low regarding the diversity of corpus information as well as the number of ontological categories. It is generally agreed that ontology design imposes to favor reliability over cover. The obvious `NP is a NP` pattern is usually disregarded as too imprecise. In the sentence "the key regulator is an example of…" it would lead to interpret the key regulator as an instance of example.

**Meronymy or part-of relation.** Meronymy has not been as much studied as hyponymy. (Berland and Charniak 1997) however adapt the above approach to find parts of physical objects in very large corpora. Their acquisition method is designed for ontology enhancement. They propose five patterns such as
`{N|Nplural}'s POSSESSIVE {N|Nplural}`
where the first N is interpreted as the whole and the second one as the part. Morphological constraints rule out quality words (words ending with –ness, -ity…), which are not supposed to refer to physical objects. Phrases like "…basements in/of buildings…", "basement in|of a building", "basement's building" are covered by the five patterns proposed. Due to these weakly constrained patterns, many potential meronyms are extracted. They are statistically ordered and proposed to an expert for validation. Results show 55% of precision among the first 50 part-whole pairs, which is quite low.

(Hishiki *et al.* 1998) proposes patterns relying on partitive verbs for biological literature: `NP consist of NP` and `NP be ... part of NP` as in "sigE is part of an operon" or in " the gerE locus consists of one gene". This work raises the same evaluation problem as the previous one.

**Synonymy.** The same approach has been experimented to detect synonymy relations in corpora. Reformulation and abbreviation patterns have been proposed (Pearson 1998): *i.e., e.g. known as, called*. (Hishiki *et al.* 1998) suggests that "termed" "designated as" and parenthesizing denote synonymy: in the sequence `NP (NP)`, the NPs are considered as synonymous, like in "spoIIIG (sigma G)".

However, the productivity of these patterns is highly dependent on the corpus (Hamon and Nazarenko 2001). For instance, in biology, parentheses do not only denote synonymy or typographic variations as in "sigma-D (sigma D)" or in,

"chloramphenicol acetyltransferase (CAT)". They may also introduce an instance as in "a small DNA segment (157 bp)", a reference, as in "in an earlier study Piggot (J. Bacteriol)" or simply, the species of interest as in "spoIIG operon from either *B. subtilis* (spoIIG (Bs)) or *C. acetobutylicum* (spoIIG (Ca))". Overall, synonymy extraction patterns are not as reliable as for hyponymy, because extraction patterns capture syntagmatic information whereas synonymy is a paradigmatic relation[5].

### *4.2.2 Model-specific relations*

A wide range of domain specific relations are examined in IE works. Elementary relations can be interpreted as attributes of a given object class. The attributes age, name, phone number, parent, birth place can be associated to a person (Embley *et al.* 1998). Various relations can hold between objects or events: from semantic roles, such as agent or patient roles, to more complex ones such as the symptom relation in the medical domain or the interaction between biological entities in genomics.

Extracting relations between entities helps to populate a database. However, extracting a relation in isolation is usually not sufficient for ontology design. The elementary relation must be structured in more complex schemata (Embley *et al.* 1998; Aone and Ramos-Santacruz 2000). For instance, in functional genomics, one of the most popular IE task aims at building enzymes and metabolic pathways, or regulation networks that can be considered as specific ontologies. Such networks are described by complex graphs of interactions between genes, proteins and environmental factors such as drugs or stress. The ontological result of the extraction should represent at least the entities, their reactions, their properties and, at a higher level, feedback cycles. Single elementary and binary relations between entities are independently extracted by IE methods. The integration of these elementary relations into the ontology highly depends on the biological model represented in the ontology and on the other extracted facts. Few works address this integration question. As shown in Sect. 3, the improvement of the ontology by IE simply comes to add new instances of the interaction relation in most of the cases. For instance, with the semantic roles associated to *repress* (Agent(Repress, Protein) and Target(Repress, Gene)), the repress relation can be enriched by new instances. "SpoIIID represses spoVD transcription" yields Agent(Repress, SpoIIID) and Target(Repress, spoVD) (Roux *et al.* 2000). Other works such as (Ng and Wong 1999) aim at providing a user-friendly interface to facilitate the interpretation of the elementary results by the biologist.

On the whole, although useful, pattern-based relation acquisition cannot be the main knowledge source for ontology design. The best results in precision are obtained in hyponymy and specific relation extractions. Some reasons can be invoked. The variation in phrasing is difficult to capture and this affects the recall quality. General patterns must rely on grammatical words or construct (like prepo-sitions) which are semantically vague. This affects the precision. More fundamentally, the linguistically-based model cannot be directly mapped onto an ontology (Bouaud *et al.* 1996; Gangemi *et al.* 2001). Hyponymy between polysemic terms cannot be considered as a transitive relation; metonymy phenomena are conceptual shortcuts, language makes the confusion between the roles and the entities that hold the roles. The use of IE relation extraction techniques must therefore be restricted to the complementation and tuning of an existing ontology and any extracted information must be further interpreted in ontological terms.

---

[5] Along the paradigmatic axis, the terms can substitute to each other; along the syntagmatic axis, terms rather tend to combine.

## 5 Conclusion

As illustrated in this chapter, the IE research related to the ontology is abundant, multiple and mainly applied. Many systems, approaches, algorithms and evaluations on quite basic applications are reported. At this stage, the main goal is more to develop systems that get a better precision and recall than making explicit and defending a given general approach against others. The influence of statistics on NLP, the influence of MUC on IE and the cost of ontological processing partially explain this. We rather interpret the *quasi* absence of clear direction and modeling attempt by the novelty of the IE field. The simplest tasks are solved first (*e.g.* named entity recognition). IE methods for interpreting the lowest text levels are now well established. This maturity and the growing needs for real applications will draw the field towards a stronger involvement of the ontological knowledge.

Difficult and unexplored questions dealing with the discrepancy between what the text is about, the exogenous lexicon and ontology should be investigated. This gap may not be only due to representation languages, to divergent generality levels and incompleteness of the knowledge sources, which have been tackled by the revision field, but also to divergent text genres, points of view and underlying problem-solving tasks. IE driven by the ontology and integration of the extracted knowledge in the ontology will not be properly done without appropriate answers to these questions.